\documentclass[letterpaper]{article} 
\usepackage{aaai2026}  
\usepackage{times}  
\usepackage{helvet}  
\usepackage{courier}  
\usepackage[hyphens]{url}  
\usepackage{graphicx} 
\usepackage{natbib}  
\usepackage{caption} 
\usepackage{algorithmic}
\usepackage[linesnumbered,ruled,vlined]{algorithm2e}
\usepackage{newfloat}
\usepackage{listings}
\DeclareCaptionStyle{ruled}{labelfont=normalfont,labelsep=colon,strut=off} 
\usepackage[dvipsnames]{xcolor}

\usepackage{tikz}
\usetikzlibrary{shapes.arrows,shapes.geometric, arrows, shadows,positioning}
\usepackage{pgfplots}
\pgfplotsset{compat=1.18}

\usepackage{xcolor}
\definecolor{ToDoColor}{rgb}{0,0.16,0.90} 
\definecolor{OutlineColor}{rgb}{0.2,0.8,0.2} 
\definecolor{CommentColor}{rgb}{0.90,0.16,0} 

\usepackage{soul} 
\usepackage{comment}
\usepackage{graphicx}
\usepackage{diagbox} 
\usepackage{multirow} 
\usepackage{csvsimple} 
\usepackage{amsmath}
\usepackage{anyfontsize}
\usepackage{booktabs}

\newcommand{\pddlp}{\textsc{pddl+}}

\newcommand{\pluseq}{\mathrel{{+}{=}}}

\title{Personalized Medication Planning via Direct Domain Modeling and LLM-Generated Heuristics}
\author {
    Yonatan Vernik\textsuperscript{\rm 1},
    Alexander Tuisov\textsuperscript{\rm 2},
    David Izhaki\textsuperscript{\rm 1},
    Hana Weitman\textsuperscript{\rm 1},
    Gal A. Kaminka\textsuperscript{\rm 1},
    Alexander Shleyfman\textsuperscript{\rm 1}
}
\affiliations {
    \textsuperscript{\rm 1}Computer Science Department, Bar-Ilan University\\
    \textsuperscript{\rm 2}Independent Researcher\\
   \{yonatanw55, queldelan, davidizhakinew\}@gmail.com,
   \{weitman, galk\}@cs.biu.ac.il, alexander.shleyfman@biu.ac.il
}

\begin{document}
\maketitle
\sloppy

\begin{abstract}
Personalized medication planning involves selecting medications and determining a dosing schedule to achieve medical goals specific to each individual patient.
Previous work successfully demonstrated that automated planners, using general domain-independent heuristics, are able to generate personalized treatments, when the domain and problems are modeled using a general domain description language (\pddlp).
Unfortunately, this process was limited in practice to consider no more than seven medications. In clinical terms, this is a non-starter.
In this paper, we explore the use of automatically-generated domain- and problem-specific heuristics to be used with general search,  as a method of scaling up medication planning to levels allowing closer work with clinicians.
Specifically, we specify the domain programmatically (specifying an initial state and a successor generation procedure), and use an LLM to generate a problem specific heuristic that can be used by a fixed search algorithm (GBFS).
The results indicate dramatic improvements in coverage and planning time, scaling up the number of medications to at least 28, and bringing medication planning one step closer to practical applications.

\end{abstract}

\section{Introduction}

Personalized medication planning (PMP) is a recent and challenging domain in automated planning~\cite{alaboud2019personalized,alon-etal:ecai-2024}. PMP involves selecting medications (drugs) and scheduling dosages to achieve patient-specific medical goals, while accounting for individual health constraints and behavioral patterns. The planner,therefore, must consider drug safety, pharmacological interactions, physiological variability, sequential and parallel treatments, repeated administrations, and drug combinations. The resulting plan determines \textit{what} drugs to administer, \textit{when}, and at \textit{what dosage}. 

Previous work successfully demonstrated that using domain-independent heuristics, automated planners are able to generate personalized medication plans.
\citet{alaboud2019personalized} used \pddlp~\cite{fox2006modelling} to model problems where
pain relief levels are maintained throughout patient-specific activities. \citet{alon-etal:ecai-2024} describe clinical goal achievement problems, involving multiple drugs and advanced pharmacological processes (e.g., effects and interactions), informed by published clinical data.

These early successes also reveal key factors hindering practical and research advances. 
While individual patient characteristic may make a specific planning problem easier or harder to solve,
%
the \emph{number of medications from which the planner selects is critical}. 
For each possible action at time $t$---a drug, with some dosage, potentially administered---the planner must not only simulates the durative, non-linear effects of the pharmacological processes, but must also consider lingering effects from earlier actions.  Interactions and combinations of multiple drugs are similarly complex.   As the number of potential drugs increases, planning becomes impractical. In practice, published state of the art methods handled seven (7) drugs at best~\cite{alon-etal:ecai-2024}.
We take a novel approach to scaling up the number of drugs by shifting  from domain-independent heuristics to automatically generated, \textit{problem-specific} heuristics, each generated and intended for exactly one specific patient. 
Maintaining the importance of general search procedures, we argue that that automatic generation of tailored problem-specific heuristics allows a better, nuanced view of the tradeoff between generality and specialization.  
Crucially, this enables scaling PMP to clinically meaningful sizes, facilitating future integration with real-world workflows.

Specifically, we follow the flexible planning approach of \citet{tuisov-etal:corr-2025}, where the planning problem is defined programmatically, and problem-specific heuristics—expressed as code—are automatically generated by large language models (LLMs) using textual and programmatic descriptions of the domain and problem. The initial patient state (including physiological parameters and drug levels), the successor function (modeling pharmacokinetics and pharmacodynamics), and the goal test (capturing therapeutic outcomes and safety constraints) are all defined programmatically. A general search algorithm (Greedy Best First Search, GBFS) is used to find satisficing solutions, guided by the generated heuristics.

We empirically evaluated the automated problem-specific planning approach in extensive experiments, using the most challenging PMP problem sets tested by~\citet{alon-etal:ecai-2024}---and then scaling them by multiplying the number of medications by four (inserting minor variations between medications).
The results indicate dramatic improvements in coverage and planning time, successfully considering up to 28 drugs in finding solutions. Critically, we observe a significant reduction in the time required to find a viable treatment plan. This represents a major step towards the future use of personalized medication planning as a clinical decision support system, potentially leading to more effective, safer, and individualized patient care.

\section{Background and Motivation}
Personalized medication planning is an area within the field of personalized treatment planning, rising in popularity, that seeks to tailor therapeutic interventions to the unique needs and medical constraints of individual patients.
Other personalized treatment planning areas include personalized radiation therapy planning~\cite{wang19,chow22,jones23},
generation of treatment protocols for patients with multiple comorbidities~\cite{wilk13,piovesan16,riano13,piovesan15,sanchez13,fdez19,mitplan21},
coordinating of healthcare providers and caregivers~\cite{amir15,amir16},
and chemical therapy planning~\cite{ferrer13}.

Broadly, the computation of medication plans is greatly affected by the number of different medications considered in the planning process, and the complexity of the pharmacological process models that are used by the planner to simulate the effects of actions.  We introduce below a brief overview of these processes, before discussing specific PMP approaches.  For a fuller explanation (for AI audience), please see~\cite{alaboud2019personalized,alon-etal:ecai-2024}.

\subsection{Pharmacokinetics and Pharmacodynamics}
For each potential drug, the planner must simulate two key pharmacological processes, to allow it
to infer the effects (in the sense of action effects) of the action of administrating a drug at some given time $t$.

\vspace{6pt}\noindent{\bf Pharmacokinetic (PK) Models} describe a drug's \emph{biodistribution trajectory}: its concentration at various \emph{biosites} (organs and tissues) over time, from the moment it is administered, until it is eliminated from the body (e.g., via urine).  PK models range from simple exponential decay models with 1 to 3 compartments~\cite{hull1979pharmacokinetics,toutain2004plasma} to  intricate models that consider multiple separate biophysical kinetic processes~\cite{gerlowski1983physiologically}. Alternatively, the biodistribution trajectory can also be represented explicitly  via interpolated curves, measuring the concentration of the drug in one or more biosites, at different times. This is the approach we take here.


\noindent{\bf Pharmacodynamic (PD) Models} describe the effects of the drug on the body. It  describes the interaction of drugs with target tissues. Specifically, PD models describe the magnitude and duration of the drug's effect as a function of its concentration (which changes continuously due to pharmacokinetic activity).
We follow~\citet{alon-etal:ecai-2024} in using the familiar  \emph{direct action} model~\cite{wright2011understanding} that predicts drug effect (and their magnitude) given a current biosite concentration, based on $E_{max}$ (the maximum effect a drug can produce at the biosite) and $EC_{50}$ (the concentration that produces 50\% of the maximum effect).

\noindent{\bf PK and PD models are needed for every drug}. The planner uses them to simulate the state-change effects of the action of administering the drug, so these predicted state trajectories can be reasoned about as part of the planning process.
The PK model predicts the concentration over time, and the PD model then computes the effect size at any point.
The models themselves may change per patient, and so ideally should be specified as part of the planning problem.
The combined information is the basis for critical issues is personalized medication planning. General and individual patient health and safety constraints can be specified over either drug concentration of effects. Similarly, predicted drug interactions (synergistic or antagonistic) can be computed based on the predicted concentrations and effects.

\vspace{-4pt}\subsection{Previous Work in Medication Planning}
\citet{alaboud2019personalized} introduced an early form of PMP, using \pddlp~\cite{fox2006modelling} to model medication problems involving \textit{a single} medication, whose level is to be maintained in the patient over time. Planning was carried out in off-the-shelf numeric planners.
For simplicity, a single bio-site (representing the whole body) was used, with a simple PK model often favored by clinical practitioners (exponential-decay (drug \emph{half-life}).
No PD model was used.

\citet{keps24} describe three different \pddlp~representation approaches for PMP, all allowing the planner to select from---and potentially administer---multiple drugs.
The representations share also the ability to represent non-parametric PK models, using explicit biodistribution trajectories, for multiple target biosites. Here again, no PD model was utilized.
In practice, problems with more than \textit{three medications} were not solved within reasonable time. 

The best-performing representation from the above was used by \citet{alon-etal:ecai-2024} as the basis, adding the direct action model as a PD model, and accounting for additive drug interactions (described later). The PK and PD data were taken from published clinical sources~\cite{akhtar2019comparative,kumar2023nanoparticle}. \citet{alon-etal:ecai-2024} report that the number of medications considered by the planner in any one problem was kept at 7,
as even relatively simple problems with more drug types could not be solved in reasonable time.

Scaling up PMP (in the number of medications considered) is vital to advancing this area.
We revisit the traditional tradeoff between domain-independent and problem-dependent solvers.
We maintain the general heuristic search algorithm, but use if with problem-specific heuristics, \emph{generated automatically by LLMs} from descriptions of the problem and the overall search mechanism.
The novel ability of LLMs to generate problem-specific heuristics is an opportunity to explore a novel nuanced approach to domain-independent automated planning:
\emph{domain-independent problem-specific planning}.

\vspace{-4pt}\subsection{LLM-Generated Heuristics and Planning}
This paper touches on several promising research directions as to the use of LLMs in planning.
Recent surveys distinguish two general models~\cite{tantakoun-etal:corr-2025,aghzal2025surveylargelanguagemodels}: \textit{LLMs as planners}, and \emph{LLM as formalizers}. The former refers
to systems where an LLM, wrapped in a supportive control loop, generates a plan.
The latter refers to systems where an LLM is used to formalize a planning problem, that is then solved using a traditional heuristic search process (with its own domain-independent heuristics).
The approach we take differs from both: it uses a general heuristic search algorithm,
but the heuristics for it are generated from a problem description.


\emph{Thought of Search}~\cite{tos:nips-2024} uses LLMs to generate symbolic search components (successor functions and goal tests), that are then used by a breadth-first search process to solve relatively small, finite search problems. We extend this direction, by employing heuristic search algorithm and use the LLM to generate heuristics. This is necessary as medication planning problems have far greater complexity.


\citet{correa2025classical} present a related approach, where classical planning tasks are given in PDDL, and are translated to a programmatic representation (in Python), for which heuristics are generated. In contrast, we assume a programmatic
representation of medication planning problems, that require the more advanced \pddlp~ representation.  This potentially allows
us greater representational flexibility, but at the cost of representing problems programmatically. The choice of the programming
language (Python, vs. Rust in our case) is also significant. We chose Rust due to the guarantees its static analysis provides.
This eases the task of validating the generated heuristics.

There exists a line of research on LLM-created or LLM-tuned heuristics in other combinatorial optimization problems~\cite{liu2024evolution,ye2024reevo}.
In these works, LLMs are used in conjunction with evolutionary algorithms to create heuristics that guide the optimization process.

\section{Medication Planning Problems Definition} We briefly describe the representation of the General Medication Planning (GMP) problem as defined by \citet{alon-etal:ecai-2024}. A GMP is a tuple $\langle \mathcal{M}, \mathcal{B}, \mathcal{P}, p^0, \mathcal{C}, \mathcal{G}\rangle$. $\mathcal{M}$ is the set of drugs that can be administered to the patient, $\mathcal{B}$ is the set of the target bio-sites of the patient, $\mathcal{P}$ is the set of properties in each bio-site, $p^0$ is the initial state of the patient (i.e., initial property values in each  biosite), $\mathcal{C}$ is the set of constraints on the properties, and $\mathcal{G}$ is the goal description: a set of conditions on the values of specific properties.  

Only one action template exists: $adm(m,d,t)$, which represents the administration of drug of type $m$, at dosage $d$, at time $t$. For notational brevity, we sometimes  use $d(m)$ to denote the medicine and dosage together. A solution plan is a sequence of applicable actions starting from $p^0$, leading to a state satisfying $\mathcal{G}$, without violating any constraints in $\mathcal{C}$ throughout the state trajectory. In reality, the sequence may be in fact be only partially-ordered, allowing more than a single drug type to be administered in parallel. However, in existing GMP implementations, only totally ordered sequences of administrations are considered. 

\paragraph{Biochemical Properties as Variables} Each bio-site $b \in \mathcal{B}$ is characterized by a set $\mathcal{P}$ of biochemical properties. The value of each property $p \in \mathcal{P}$ in a bio-site $b$ at any given time $t$ is represented by a numeric fluent $b[p](t)$, measured in standard units.  The initial state of the patient is given as a vector of values $b[p^0] := b[p](0)$ (for each $p\in\mathcal{P}$ and  $b\in\mathcal{B}$) defines the initial values of all fluents. For drug concentration properties, initial values are typically zero.

\paragraph{Drug Effects (1): Pharmacokinetics}
After administration, the biodistribution trajectory of the drug is given by the PK model.  For each drug $m$ and bio-site $b$, there is a property $b[m]$ representing the drug concentration. We model biodistribution using explicitly represented trajectories. For a single administration of drug $m$ at time $t_0^m$, the concentration $b[m]$ at time $t \geq t_0(m)$ is given by the formula~\cite{wright2011understanding,felmlee2012mechanism}:
$$b[m] := g(b,m,t-t_0^m) \cdot d(m),$$
where $g(b,m,\Delta t)$ is the value of the biodistribution trajectory of drug $m$ in bio-site $b$ at time $\Delta t = t-t_0(m)$, as a percentage of the initial dosage per unit mass. To handle repeated administrations of the same drug type we ensure that each administration action $adm(m_i, d_i, t)$  has not been administered in the same time step $t$.


\paragraph{Drug Effects (2): Pharmacodynamics}
PD models are used to predict how drug concentration affects biochemical properties. The effects are updated at each time step. We use the direct action model~\cite{wright2011understanding}, where the effect on a property $p$ in bio-site $b$, due to a drug $m$ is given by:
$$b[p] \pluseq E_{max}(m,b,p) \cdot \frac{b[m]}{b[m] + EC_{50}(m,b,p)},$$
where $E_{max}(m,b,p)$ is the maximum effect and $EC_{50}(m,b,p)$ is the concentration at half-maximal effect. Both parameters are specific to the drug, bio-site, and property. These updates are triggered at each time step, using the current total concentration $b[m]$. To allow for the summation of effects from different drugs on the same property, we reset all non-drug concentration property values to zero at the beginning of each time step. This models \emph{Loewe-additivity}, a common reference model of drug interactions~\cite{berenbaum77,wright2011understanding}. 

\paragraph{Constraints and Goals} The safety constraints $\mathcal{C}$ are checked in every state. Any state (set of all properties and their values at time $t$) that violates a safety constraint is declared to be a dead-end.  These constraints can also model drug-drug interactions, e.g., constraining one property's value to be lower than a threshold, when the value of another property is higher than some other threshold. 

The goal description $\mathcal{G}$ comprises three parts. First, it specifies target levels for relevant properties in therapeutic bio-sites, each of which must have been reached sometime during the plan. Second, it requires that all safety constraints remain satisfied throughout the entire plan. Third, it requires that all administered drugs are cleared from the patient's body. This ensures that the therapeutic effect is achieved safely and maintained until the treatment is complete.

\paragraph{Personalization} The GMP framework is intended for personalization in several ways. The domain description (as published by~\citet{alon-etal:ecai-2024}) contains the generic drug administration action $adm(m,d,t)$,  effect computation machinery and constraint checks, and  a list of all potential drugs. However, it is the problem description---the patient's medication planning problem---that creates the personalization, including the actual list of drugs to be considered for this patient. The description also includes the patient-specific initial state (property values), the personalized biodistribution trajectories and PD parameters ($E_{max}, EC_{50}$), personal safety constraints (e.g., based on pre-existing conditions), and therapeutic goals. This enables the generation of tailored medication plans that account for individual patient characteristics and needs.

\section{Searching with LLM-Generated Heuristics}
We begin with a brief overview of the approach, and then provide details on two key steps: generation of a problem-specific heuristic, and robust usage, given allotted resources.

\paragraph{Overview:} We follow the \emph{Search + LLM-generated heuristics} paradigm proposed by \citet{tuisov-etal:corr-2025}, and apply it to GMP as presented above, using Rust to represent the problem and heuristics programmatically. We employ Greedy Best-First Search (GBFS) with an instance-specific heuristic generated by GPT-5.1, based on the findings of \cite{tuisov-etal:corr-2025}. The key steps are:

\paragraph{1. Programmatic Representation}  State change trajectories, resulting from administering drugs, are represented as \emph{successor generator} functions in Rust. These receive as input the current state, and calculate all valid actions and resulting states.  For the specific problem, boolean \textit{goal-test} and \textit{constraint checking} functions, were written manually in Rust. Future work will examine automatic generation of the Rust code from the problem description.

\paragraph{2. Structured Problem Representation}
Given a medication planning problem, we translate it to a JSON representing the starting state, including organs, their properties, starting values, and safety limits, as well as medicines, allowed dosage sizes and amount, and their PDPK parameters. The instance JSON also contains the goal specification.\footnote{See Supplementary Materials for a JSON example.} This translation can be done automatically from a \pddlp\ representation as formulated in \cite{alon-etal:ecai-2024}.\footnote{We implemented a Python translator from \pddlp\ to JSON.}


\paragraph{3. Creating a problem-specific planner}
Finally, the Rust code is given to a LLM to generate a suitable heuristic in Rust.
The heuristic is compiled and linked with the heuristic search algorithm (GBFS), resulting in a problem-specific planner, written in Rust and compiled.

\subsection{Generating Heuristics}
We follow ~\cite{tuisov-etal:corr-2025}, but the method is briefly reiterated here. A heuristic tailored for the domain is generated using a prompt with three components:

\paragraph{\textit{Set up}} Conditioning the model to be a senior Rust engineer \cite{Anam2025PromptEA} and providing the format the resulting heuristic must follow. This component is spread across both the system prompt and the user message.\footnote{As noted in Tuisov. et al, the prompt is not well optimized. We find it sufficient for this problem.}

\paragraph{\textit{Task}} Requesting that the model generate a heuristic and that it consider the patient's medical history.

\paragraph{\textit{Context}} Providing the model with the Rust domain implementation.




Following Tuisov et al, we generate up to 10 heuristics and count a success if any one of them succeeds \cite{tuisov-etal:corr-2025}. The full prompt is included in the Supplementary Materials. 

\paragraph{An Example Heuristic} We present a high-performing heuristic by Claude Sonnet 4.5. For brevity, we present it in pseudo-code (Alg.~\ref{alg:heuristic}); the complete code is included in the Supplementary Materials: it has 342 lines, of which 39 are pure comments. Alongside the heuristic function's body, the model generated 8  utility functions.

\begin{algorithm}[htbp]
\caption{Comprehensive Heuristic for Medical Treatment Planning}
\label{alg:heuristic}
\KwIn{Medical problem \texttt{problem}, state \texttt{state}}
\KwOut{Estimated minimum timesteps to safely achieve all goals}

\SetKwFunction{FMain}{Heuristic}
\SetKwFunction{FMaxGoal}{MaxGoalTime}
\SetKwFunction{FClearNow}{ClearTime}
\SetKwFunction{FClearFuture}{ClearancePenalty}
\SetKwFunction{FSafety}{SafetyPenalty}
\SetKwProg{Fn}{Function}{:}{}

\Fn{\FMain{problem, state}}{
    \If{\texttt{state.goals\_remaining} empty}{
        \Return \FClearNow{problem, state} \tcp*{Wait for current medicines to clear}
    }

    \texttt{max\_goal\_time} $\leftarrow$ \FMaxGoal{problem, state}\;
    \texttt{clearance\_penalty} $\leftarrow$ \FClearFuture{problem, state}\;
    \texttt{safety\_penalty} $\leftarrow$ \FSafety{problem, state}\;

    \Return \texttt{max\_goal\_time} + \texttt{clearance\_penalty} + \texttt{safety\_penalty}\;
}

\vspace{0.1cm}
\Fn{\FMaxGoal{problem, state}}{
    \texttt{max\_goal\_time} $\leftarrow 0$\;
    \ForEach{((\texttt{organ}, \texttt{prop}), \texttt{req}) in \texttt{state.goals\_remaining}}{
        \texttt{cur} $\leftarrow$ current value of (\texttt{organ}, \texttt{prop}) (default $0$)\;
        \texttt{deficit} $\leftarrow \texttt{req} - \texttt{cur}$\;
        \If{\texttt{deficit} $\le 0$}{\textbf{continue}}
        
        \tcp{Pick medicine/dosage with strongest effect}
        \texttt{med} $\leftarrow$ best available medicine for (\texttt{organ}, \texttt{prop})\;
        \texttt{peak\_eff} $\leftarrow$ peak E\textsubscript{max}-based effect of \texttt{med}\;
        \texttt{cur\_contrib} $\leftarrow$ current effect of \texttt{med} in this organ/property\;
        \texttt{adj\_def} $\leftarrow \max(0, \texttt{deficit} - \texttt{cur\_contrib})$\;
        \texttt{doses} $\leftarrow$ ceil(\texttt{adj\_def}/\texttt{peak\_eff}), clipped by remaining doses\;
        
        \texttt{t\_peak} $\leftarrow$ time to peak PK effect of \texttt{med}\;
        \texttt{elim} $\leftarrow$ elimination time of \texttt{med}\;
        \texttt{spacing} $\leftarrow \max(1, \texttt{elim}/2)$ \tcp*{safe inter-dose gap}
        \texttt{t\_goal} $\leftarrow \texttt{t\_peak} + (\texttt{doses}-1)\cdot\texttt{spacing}$\;
        
        \texttt{max\_goal\_time} $\leftarrow \max(\texttt{max\_goal\_time}, \texttt{t\_goal})$\;
    }
    \Return \texttt{max\_goal\_time}\;
}

\vspace{0.1cm}
\Fn{\FClearFuture{problem, state}}{
    \tcp{Like \FMaxGoal, but only account time of a single strongest injection per goal}
    \Return MaxDecayTimeOfBestSingleInjection(\texttt{problem}, \texttt{state.goals\_remaining})\;
}

\vspace{0.1cm}
\Fn{\FSafety{problem, state}}{
    \tcp{Linear penalty when within $20\%$ of min/max safety bounds}
    \Return BoundaryMarginPenalty(\texttt{problem.property\_constraints}, \texttt{state.organ\_properties})\;
}
\end{algorithm}

The heuristic first checks if all organs reached their desired properties, and if so exits early returning the wait time necessary for the medicines to leave the body. Otherwise, it returns the sum of three components: $max\_goal\_time$, $clearance\_penalty$, and $safety\_penalty$.

$max\_goal\_time$ takes maximum over (organ, property) pairs still required for how long each would take. For each, it determines which medicine and dosage would yield the strongest effect, then how long it would take to administer the requisite number of dosage for it to fulfill that property goal including waiting for the effect, assuming for safety that the administrations are separated by a duration of half the medicine's elimination time. Note that this does not imply that the solution must wait this long between injections.

$clearance\_penalty$ is a variation on $max\_goal\_time$ which instead of trying to account for the length of a full treatment regimen only considers the time for the single most potent injection available.

$safety\_penalty$ penalizes when the state is within 20\% of the safe limits, the penalty increasing linearly as the gap shrinks.

\subsection{Robust Usage of the Approach}
LLM-generated heuristics may not always be informative, syntactically valid, or memory-efficient. To address these challenges, we propose a robust meta-algorithm that operate under fixed resource constraints: a given time budget and a given memory limit.

\paragraph{Algorithm 2: Retry on Failure or Out-of-Memory}
\begin{enumerate}
    \item Query the LLM for a heuristic and attempt to compile it. If compilation fails, request a new heuristic and retry.
    \item Once compilation succeeds, run GBFS with the compiled heuristic.
    \item If the search process fails due to an out-of-memory (OOM) error, return to Step 1.
    \item If the overall algorithm, including generations and search, exceeds the time limit, terminate with failure. Otherwise, report the solution.
\end{enumerate}

These strategies aim to maximize the utility of LLM-generated heuristics while providing resilience to failure modes such as compilation errors and excessive memory consumption. We note that generating heuristics takes the lion's share of the time in the algorithm (see \ref{tab:costs-table}). Inference throughput tends to increase as batch size increases \cite{garcia-etal:iccc-2025}, so one could further optimize this by generating many heuristics simultaneously, however improving parallelization is out of scope for this work.

\section{Experimental Evaluation}
To evaluate the LLM-generated problem-specific approach, we implemented the PK/PD simulation model in Rust, adhering to the same parameterization used in prior work~\cite{alon-etal:ecai-2024}. 
We contrast its results with results achieved by state-of-the-art domain-independent heuristic search methods, on scaled-up versions (in number of medicines) of the most challenging benchmark sets.
The experimental settings and results are described below.

\paragraph{Problem-specific, LLM-generated planner} We used the LLM-generated heuristics as described. As a retry strategy, we used Algorithm 3 (above) with a time-slice of 100s.

\paragraph{Domain-independent heuristic planners}  As a basis for benchmark domain-independent heuristic planner, we utilized the ENHSP-20 numeric planner~\cite{scala-et-al:jair-2020}, which was used also by~\citet{alon-etal:ecai-2024}. Per personal communication with the ENHSP author,\textit{ we used the best results from either of two compiled binaries of ENHSP-20}: One from the ENHSP-20 website (\url{https://sites.google.com/view/enhsp/}), and one  from the latest branch (\url{https://github.com/hstairs/enhsp/tree/enhsp-20}).

ENHSP-20 has implementations of many domain-independent heuristics for satisfycing and optimal search. We compared against \emph{all} that could work with the original \pddlp~ problem sets. We note that while the \emph{sat-hmd} and the \emph{novelty-based heuristics} proposed in \citet{chen-thiebaux:socs-2024} are theoretically applicable, their implementation was declared to be incompatible with the \pddlp~ problem descriptions~\cite{chen-personal}.


\paragraph{Resource and Time Limits} Each problem instance was run with a time budget of 600 seconds and an 16GB memory cap. Experiments were run on an dual-CPU machine (Intel Xeon CPU E5-2630 v4 @ 2.20GHz) with 256GB RAM. Tests were conducted in parallel using 20 hyperthreaded cores (a total of 40 logical cores). No more than 20 tests were run in parallel. 

\paragraph{Benchmark Problem Sets} We used all seven benchmark domains introduced by Alon et al.~\shortcite{alon-etal:ecai-2024}. Each of these has 37 problem instances.  These are grounded in a pharmacokinetics/pharmacodynamics (PK/PD) model using data from a large-scale database that captures biodistribution trajectories for over 200 nanoparticle-based drugs across up to 15 bio-sites in mice or rats~\cite{kumar2023nanoparticle}. 
Pharmacodynamic data was drawn from prior studies~\cite{lezama2021time,roman2021combination,santos2023role,alon-etal:ecai-2024}. The baseline domain uses original values. The other domains tweak these to generate harder and easier sets by manipulating safety constraints and the duration of pharmacological processes. For details on dataset construction and domain modeling, refer to Section 4 in~\cite{alon-etal:ecai-2024}.

To mitigate the stochastic variability inherent to LLM-based heuristic generation, we repeated each experiment with three different LLM generation seeds. We report coverage using the minimum, median, and maximum number of problems solved per domain across these seeds.

\paragraph{Baseline: Domain-Independent Heuristics}
We first establish a baseline, allowing us to identify the toughest benchmarks set for domain-independent heuristics. Table~\ref{tab:classical} presents the results of domain-independent heuristics on all original benchmark problems. We report results only for configurations (algorithm+domain-independent heuristic) that managed to solve at least one problem (exception: \texttt{sat-add}, not shown, solved 2 problems in total).

\begin{table}[ht]
\centering
\small
\setlength{\tabcolsep}{2pt}
\begin{tabular}{@{}l|rrrrrrrrrr@{}}
\toprule
Domain (37 problems) & opt-blind & opt-hmax & opt-hrmax & sat-aibr \\
\midrule
baseline                        & 3  & 37 & 34 &  3 \\
constraints-loose                   & 3  & 37 & 33 &  3 \\
constraints-tight                & 3  & 19 & 14 &  1 \\
time-shrinkage-factor2            & 3  & 37 & 37 &  8 \\
time-shrinkage-factor4              & 12 & 37 & 37 & 16 \\
time-stretching-factor2           & 3  & 34 & 21 &  0 \\
time-stretching-factor4             & 0  & 21 & 13 &  0 \\
\midrule
Total                                     & 27 & 222 & 189 & 31 \\
\bottomrule
\end{tabular}
\caption{coverage of best configuration of ENHSP-20 on the domains by Alon et al.~\shortcite{alon-etal:ecai-2024}.
}
\label{tab:classical}
\end{table}

\subsection{Scaling the Domains}
We now focus our experiments on the most challenging variants of the baseline medication planning domains introduced by Alon et al.~\shortcite{alon-etal:ecai-2024}: \textit{Tight Constraints (Tight)} and \textit{Time Stretching X4 (Stretching X4)}. 

The \textit{Tight} setting reduces the permissible safety margins, placing the safety threshold just above the required therapeutic level in each bio-site. As a result, many otherwise valid plans are rendered unsafe, demanding highly precise treatment schedules. The \textit{Stretching X4} setting extends the time horizon by a factor of four, meaning the planner must reason over longer treatment durations. This increases the difficulty in two ways: (1) the temporal span of dependencies between actions grows, and (2) the length of valid plans becomes four times longer, significantly expanding the search space. 

While these settings offer meaningful challenges, the original domains remain relatively small and do not fully capture the complexity of realistic treatment planning. We therefore created scaled-up versions by multiplying the number of available medications in each domain by a factor of 4. This increases the number of action options in each step, as every drug has its own PK/PD parameters and potential interactions. From a planning perspective, this drastically increases the branching factor; from an LLM perspective, it introduces substantial noise by burying relevant causal structure among irrelevant details. 

Thus from this point we will be evaluating on four \pddlp\ benchmark domains, each containing 37 problem instances: the original \textit{Tight}, \textit{Stretching X4}, and their counterparts with 4$\times$ the number of medications, \emph{Tight (4XMeds} and \emph{StretchingX4 (4XMeds)}.

Table~\ref{tab:llm-vs-enhsp-updated} compares our method against the best-performing heuristics from ENHSP-20. Coverage is measured as the number of problems solved out of 37. In all four domains, our method significantly outperforms both opt-max and opt-hrmax, especially in the most difficult settings. To clarify, the use of any of the domain-independent heuristics built into ENSHP did not yield any solutions.

\begin{table}[!t]
\centering
\footnotesize
\setlength{\tabcolsep}{3pt}
\begin{tabular}{@{}l|rrrr@{}}
\toprule
\textbf{Problem Setting} & \textbf{opt-hmax} & \textbf{opt-hrmax} & \textbf{GPT-5.1} & \textbf{CS-4.5} \\
\midrule
{\scriptsize Tight (37)}                     & 14 & 10 & 21 & 28 \\
{\scriptsize StretchingX4 (37)}              & 17 &  6 & 37 & 37 \\
{\scriptsize Tight (4xMeds.) (37)}           &  8 &  4 &  0 & 29 \\
{\scriptsize StretchingX4 (4xMeds.) (37)}    &  2 &  0 & 30 & 36 \\
\midrule
{\scriptsize Total (148)}                    & 41 & 20 & 114 & 130 \\
\bottomrule
\end{tabular}
\caption{\label{tab:llm-vs-enhsp-updated}
Coverage of ENHSP-20's most performant heuristics vs. GPT-5.1 and Claude Sonnet-4.5 (CS-4.5), on the hardest domains and their scalings. LLMs were run by generating a heuristic and running it, then repeating if it fails until either a heuristic succeeds or the resources are exhausted (600 seconds, 16GB RAM).}
\end{table}


The results show that generating multiple heuristics and refining them improves coverage across all domains. In particular, the combined unrefined + refined strategy delivers the best performance. Even with very limited search time (10 seconds), the refined heuristics remain competitive.

Note that the unrefined approach allows us to generate multiple heuristics for the same domain in a single batch, which can then be reused across different runs. This not only diversifies the heuristic set but also opens up the possibility of combining multiple heuristics to improve performance. We leave this direction for future work.

\begin{figure}[tp]
    \centering
    \includegraphics[width=0.48\textwidth]{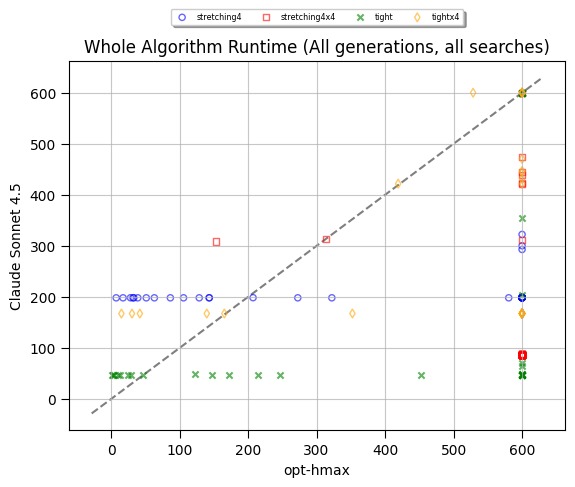}
    \caption{\label{fig:runtime} Per-instance comparison of the runtime of opt-hmax vs. Our Algorithm on Claude Sonnet 4.5. Points below the diagonal favor our approach. We observe a pattern where each domain's instances are approximately delineated into horizontal lines depending on which \#generation solved a given instance.}
\end{figure}

\paragraph{Runtime Comparison} Figure~\ref{fig:runtime} compares the runtime of our method to opt-hmax on a per-instance basis. Our method consistently solves more problems in less time, especially in the 4$\times$-medications settings, where classical planners struggle with time or memory limits.



\paragraph{Summary} Across all domains, our LLM-generated heuristics demonstrate clear advantages over traditional ones. They solve more problems overall, especially in larger and more challenging domains, and they do so within realistic computational budgets. The method is robust to noise and occasional failures, and scales well with problem size. When a generated heuristic fails to compile or performs poorly, a retry strategy proves effective: several heuristics can be generated and evaluated within a fixed time budget (e.g., 100 seconds), giving the system multiple opportunities to produce high-quality guidance.

In the most challenging domain (StretchingX4 with 4×Medications), the best ENHSP-20 heuristic solves only 2 problems, while our method solves up to 28. This gap highlights how quickly classical heuristic approaches degrade as domain complexity increases, whereas our LLM-generated heuristics remain effective. Even when taking the minimum number of problems solved across repeated runs, our method still solves 80 problems in total—compared to just 41 by opt-hmax. Notably, the minimum performance of our approach exceeds that of opt-hmax in every individual domain.

\begin{figure*}[ht]
    \centering
    \includegraphics[width=1.0\textwidth]{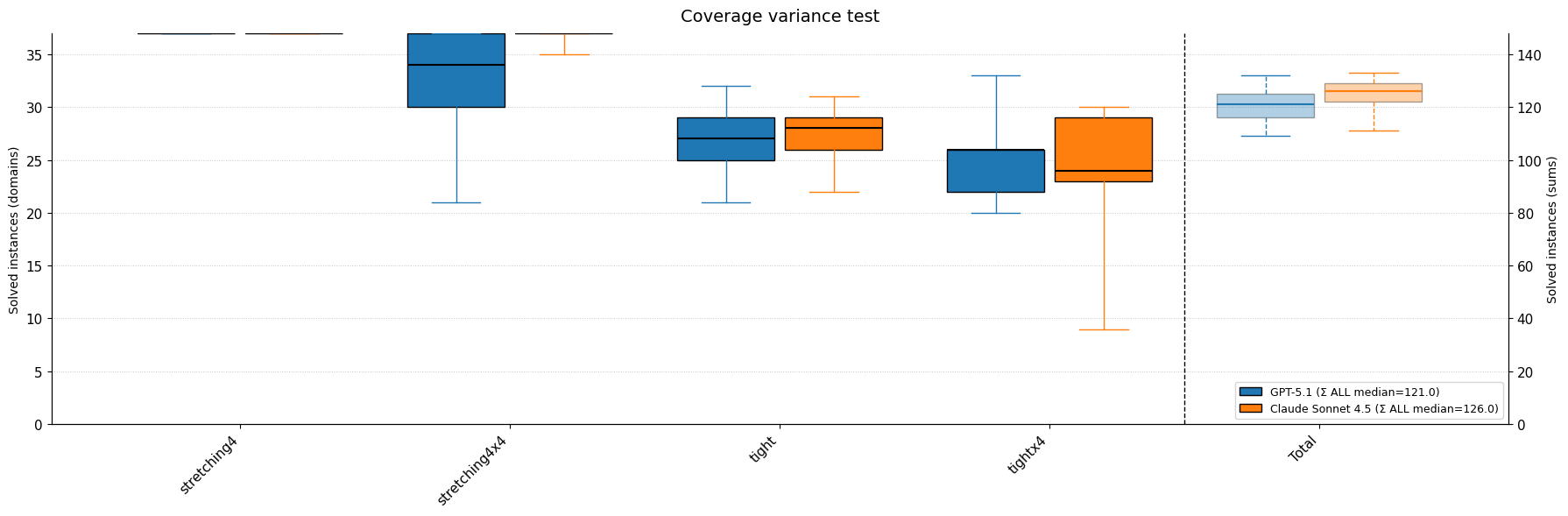}
    \caption{For medical applications it is important to have low variance. As a variance analysis we generated 30 heuristics in each domain with each model, and performed a Monte-Carlo simulation of our algorithm for 1000 iterations of sampling heuristic order, taking the coverage per-domain and overall each time. The boxes represent the first and third quartiles, the thick lines the median, and the whiskers the 1st and 99th percentiles. Although not perfect, especially in Tight(4xMeds), the method is fairly consistent, we still consistently outperform existing methods significantly, both by-domain and overall. A mitigating factor is that thanks to the formulation of heuristic search, a failure only 'returns you to square one', and does not lead to a \textit{wrong} treatment.}
    \label{fig:variance}
\end{figure*}

\begin{table}[ht]
\centering
\begin{minipage}{0.48\textwidth}
\centering
\begin{tabular}{lcc}
\toprule
\textbf{Metric} & \textbf{GPT-5.1} & \textbf{Sonnet 4.5} \\
\midrule
Input tokens            & 4683  & 5903   \\
Output tokens           & 4349  & 2619   \\
Reasoning tokens\footnote{For GPT-5.1, subset of output tokens. For Sonnet, they are not provided by the API}        & 1547  & -      \\
Cost (USD)              & 0.049 & 0.057  \\
Generation time (s)     & 58.7  & 43.7   \\
Run time (s)            & 46.0  & 10.5   \\
\midrule
Compilation error rate  & 0.069 & 0.575 \\
OOM rate                & 0.342 & 0.128 \\
Runtime error rate      & 0.000 & 0.002 \\
Timeout rate            & 0.016 & 0.000 \\
Success rate            & 0.573 & 0.295 \\
\bottomrule
\end{tabular}
\caption{\label{tab:costs-table}Average token usage, cost, latency, and execution outcomes for GPT-.51 and Claude Sonnet 4.5 over 120 generations each, as standalone from the algorithm. One would expect from the individual success rates that GPT-5.1 would win - at least one reason it does not is that Sonnet heuristics are more efficient - their average runtime is  shorter by a factor of $\sim$4.6, and remains shorter by a factor of $\sim$2 even if we only consider compiling heuristics from both. Combined with lower generation time, this allows Sonnet to 'fail faster' and make on average $\sim2.74\times$ more attempts.}
\end{minipage}
\end{table}

\section*{Ethics and Societal Implications}




Personalized medication planning (PMP), a form of medical decision-support, aims to tailor treatments to individuals and avoid the limitations of population-level protocols. Our work evaluates whether AI planning can achieve such fine-grained personalization. Although the simulation data are drawn from scientific publications, the research is preliminary and lacks any clinical or animal validation. Clinical use would require addressing key issues: avoiding over-reliance on automated interpretations, ensuring reliable operation in real-world conditions (e.g., precise timing), predicting long-term effects, and assessing factors necessary for safe deployment. Clinicians must be able to understand AI-generated plans without added workload, which requires clear visualization and explanation tools~\citep[see, e.g.,][]{explimed24}. These are critical directions for future research.

LLM-generated heuristics introduce additional risks. Comprehensive auditing is essential: data sources, prompt chains, model versions, and outputs must be logged immutably for reproducibility and forensic review. Pre-deployment testing must include adversarial and boundary cases that probe rare physiology, atypical PK/PD parameters, and operational edge conditions such as timing errors or missing data. Independent audits should confirm intended system behavior, and continuous monitoring is needed to detect distribution shift and emerging safety violations.

Explainability is equally crucial. Clinicians require structured summaries of each plan -- its assumptions, expected drug effects, constraints, and main uncertainty drivers. Plans should include short, interpretable traces showing why actions were selected and how they affect predicted trajectories, supported by visual timelines of concentrations, effects, and safety margins. Because uncertainty in PK/PD parameters and in model-generated heuristics can meaningfully alter outcomes, the system must present confidence levels and highlight brittle or weakly supported recommendations. This demands extensive data on the PK/PD behavior of drugs used in simulation.

Deployment adds operational risks absent in simulation. Real workflows involve irregular timing, adherence variability, incomplete data, and outages. Planners must tolerate delays, missed doses, and inconsistencies, and fall back to conservative policies when needed. Human oversight remains central: clinicians must approve all plans, and interfaces should counter automation bias. Security and privacy measures, including controlled access, encrypted logs, versioned consent, are required to limit secondary risks.

Finally, governance is needed to prevent silent system degradation. Any change to models, prompts, parameters, or generation pipelines must trigger re-validation and re-audit. Logging, monitoring, and rollback procedures must be established in advance. Given that this work is a proof of concept with no biological testing, substantial technical, clinical, and regulatory work is required before any real-world use. Addressing auditing, explainability, and deployment risks is therefore a prerequisite for responsible progress toward clinical evaluation.

\section{Conclusion}

This paper has presented a novel approach to scaling PMP \cite{alon-etal:ecai-2024} by moving away from domain-independent heuristics to automatically generated, problem-specific and domain specific heuristics provided by a LLM, coupled with a classic heuristic search algorithm (\emph{GBFS}).
We evalutated this approach on the two hardest PMP settings from prior work, \textit{Tight Constraints} and \textit{Stretching X4}, and scaled them further by multiplying the number of medications by four.

By specifying the domain programmatically and using an LLM to generate a problem-specific heuristic, we were able to dramatically improve both planning time and problem coverage (specifically, we have shown this with both GPT-5.1 and Sonnet-4.5).  While previous planners could not handle more than seven drugs, our approach successfully scaled to 28, enabling realistic planning for complex treatment plans.
Across all four domains, our method outperformed ENHSP-20 \cite{scala-et-al:jair-2020}, especially in the harder settings. In the most challenging domain (StretchingX4 with 4×Meds), we solved up to 36 problems, compared to just 2 by the best domain-independent heuristic. This demonstrates that traditional heuristics degrade quickly with complexity, while LLM-generated heuristics remain effective and robust.

This work shows that LLM-generated heuristics can significantly improve coverage and speed in PMP, a domain that is particularly challenging for automated planning. This is a step towards making automated planning more viable for clinical use, though much remains for future work on both algorithmic and clinical fronts. We note in particular that
future work should include a focus on clinical integration, interpretability tools for clinicians, and ensuring reliable and ethical deployment. 


\bibliography{aaai2026,aaaivar}
\end{document}


\maketitle

\section*{Prompt}

Below is the system prompt and prompt used in all generations and experiments. [RUST\_DOMAIN\_DEFINITION] is replaced by the domain definition as encoded in rust.

The only change from Tuisov et al. is the addition of the medical-specific line: "In your heuristic, consider the patient's medical history, the effect and safety of treatments the patient already received, and potential future treatments and their effects."

\subsection*{System Prompt}

\begin{lstlisting}[style=boxed, caption={}, basicstyle=\tiny, breaklines=true]
You are a senior Rust engineer specializing in heuristic design for forward-search planners.
- Always produce compiling, self-contained code.
- Never introduce types not present in the domain file.
- Prefer clarity, comments, and determinism over clever tricks.
- Heuristic values should decrease as we approach the goal (admissibility is optional).
- Helper functions must end with `_heuristic_helper` and live in the same block as the heuristic.
- When matching, include a default `_ =>` arm.
- Assume unit action costs unless the domain says otherwise.
- Output EXACTLY one Rust code block when asked for code.
\end{lstlisting}

\subsection*{Prompt}

\begin{lstlisting}[style=boxed, caption={}, basicstyle=\tiny, breaklines=true]
I have a Rust implementation of a forward-search problem. The main algorithm is complete; I need a strong heuristic.
Please invent an effective heuristic for this domain, then implement it in Rust.
In your heuristic, consider the patient's medical history, the effect and safety of treatments the patient already received, and potential future treatments and their effects.

Important tips/guidelines:
1) Comment your reasoning in code.
2) Be careful with numeric casts (e.g., `as f32`).
3) When using a match arm, add an emergency default arm `_ =>`.
4) If you create helper functions, end their name with `_heuristic_helper` and put them in the same code block.
5) Write the complete heuristic; do not leave placeholders. The code must be a finished product.
6) Assume action costs=1 for all actions

Write the code in the module scope (i.e. just start writing the function/helper functions, don't go into implementation scope)
The function signature must be precisely:

```rust
fn heuristic(problem: &ProblemType, state: &State) -> f64
```

(Replace `ProblemType` with the actual type from the domain.)

domain definition:
------------------
[RUST_DOMAIN_DEFINITION]
\end{lstlisting}

\section*{Example heuristic}
\begin{lstlisting}[language=Rust, style=boxed, caption={Example heuristic generated by Claude Sonnet 4.5. The heuristic is provided as-is, and we note that all comments in the code were generated by the model itself}, basicstyle=\tiny, breaklines=true]
/// Comprehensive heuristic for the medical treatment planning domain.
/// 
/// This heuristic estimates the minimum cost (timesteps) to achieve all remaining goals
/// while respecting safety constraints. It considers:
/// 1. Goals that still need to be achieved
/// 2. Current medicine levels in the body (PK state)
/// 3. Time needed for medicines to take effect and then clear
/// 4. Safety margins and constraints
/// 5. Medicine usage constraints (remaining doses available)
fn heuristic(problem: &MedicalProblem, state: &State) -> f64 {
    // If all goals are achieved, estimate time for all medicines to clear the body
    if state.goals_remaining.is_empty() {
        return estimate_clearance_time_heuristic_helper(problem, state);
    }

    // Calculate the maximum estimated time to achieve each remaining goal
    let mut max_goal_time = 0.0_f64;
    
    for ((organ, property), required_level) in &state.goals_remaining {
        let current_level = state.organ_properties
            .get(&(organ.clone(), property.clone()))
            .map(|of| of.0)
            .unwrap_or(0.0);
        
        let deficit = required_level.0 - current_level;
        
        // Find the best medicine(s) that can affect this organ-property pair
        let time_for_goal = estimate_time_to_reach_goal_heuristic_helper(
            problem,
            state,
            organ,
            property,
            deficit,
        );
        
        max_goal_time = max_goal_time.max(time_for_goal);
    }
    
    // Add estimated clearance time after achieving goals
    let clearance_penalty = estimate_future_clearance_heuristic_helper(problem, state);
    
    // Add penalty for being close to safety boundaries (encourages safer paths)
    let safety_penalty = compute_safety_penalty_heuristic_helper(problem, state);
    
    // Combine estimates
    max_goal_time + clearance_penalty + safety_penalty
}

/// Estimates the time needed for all currently active medicines to clear from the body
fn estimate_clearance_time_heuristic_helper(problem: &MedicalProblem, state: &State) -> f64 {
    let mut max_clearance = 0;
    
    for medicine in &problem.static_info.medicines {
        let elimination_time = *problem.static_info.decay_times.get(medicine).unwrap_or(&0);
        
        if let Some(doses) = state.medicine_history.get(medicine) {
            for &(dose_time, _dosage) in doses {
                let elapsed = state.timestamp - dose_time;
                let remaining = elimination_time - elapsed;
                if remaining > max_clearance {
                    max_clearance = remaining;
                }
            }
        }
    }
    
    max_clearance as f64
}

/// Estimates future clearance time based on medicines we'll likely need to use
fn estimate_future_clearance_heuristic_helper(problem: &MedicalProblem, state: &State) -> f64 {
    let mut max_future_clearance = 0.0_f64;
    
    // For each remaining goal, find which medicine would be most effective
    for ((organ, property), _required_level) in &state.goals_remaining {
        let (best_medicine, _best_emax) = find_best_medicine_for_goal_heuristic_helper(
            problem,
            organ,
            property,
        );
        
        if let Some(med) = best_medicine {
            let elimination_time = *problem.static_info.decay_times.get(&med).unwrap_or(&0);
            max_future_clearance = max_future_clearance.max(elimination_time as f64);
        }
    }
    
    max_future_clearance
}

/// Find the medicine with the highest Emax for a given organ-property pair
fn find_best_medicine_for_goal_heuristic_helper(
    problem: &MedicalProblem,
    organ: &String,
    property: &String,
) -> (Option<String>, f64) {
    let mut best_medicine: Option<String> = None;
    let mut best_emax = 0.0_f64;
    
    for medicine in &problem.static_info.medicines {
        let key = (medicine.clone(), organ.clone(), property.clone());
        if let Some(&emax_val) = problem.static_info.emax.get(&key) {
            if emax_val > best_emax {
                best_emax = emax_val;
                best_medicine = Some(medicine.clone());
            }
        }
    }
    
    (best_medicine, best_emax)
}

/// Estimates the time (in timesteps) needed to reach a specific goal deficit
fn estimate_time_to_reach_goal_heuristic_helper(
    problem: &MedicalProblem,
    state: &State,
    organ: &String,
    property: &String,
    deficit: f64,
) -> f64 {
    if deficit <= 0.0 {
        return 0.0;
    }
    
    // Find the best medicine for this organ-property combination
    let (best_medicine_opt, best_emax) = find_best_medicine_for_goal_heuristic_helper(
        problem,
        organ,
        property,
    );
    
    let best_medicine = match best_medicine_opt {
        Some(med) => med,
        None => return 1000.0, // Large penalty if no medicine can help
    };
    
    // Check if we have doses remaining for this medicine
    let doses_applied = state.medicine_doses_applied.get(&best_medicine).unwrap_or(&0);
    let max_doses = problem.static_info.usage_constraints.get(&best_medicine).unwrap_or(&i32::MAX);
    
    if doses_applied >= max_doses {
        // Can't use this medicine anymore, find next best
        return find_alternative_medicine_time_heuristic_helper(
            problem,
            state,
            organ,
            property,
            deficit,
            &best_medicine,
        );
    }
    
    // Get the maximum dosage available for this medicine
    let max_dosage = problem.static_info.dosage_sizes
        .get(&best_medicine)
        .and_then(|sizes| sizes.iter().max())
        .unwrap_or(&1);
    
    let key = (best_medicine.clone(), organ.clone(), property.clone());
    let ec50_val = problem.static_info.ec50.get(&key).unwrap_or(&1.0);
    
    // Estimate peak effect: Emax * (dosage / (dosage + EC50))
    let peak_effect = best_emax * (*max_dosage as f64 / (*max_dosage as f64 + ec50_val));
    
    if peak_effect <= 0.0 {
        return 1000.0;
    }
    
    // Estimate number of doses needed
    let doses_needed = (deficit / peak_effect).ceil();
    
    // Get PK profile to estimate time to peak
    let time_to_peak = estimate_time_to_peak_heuristic_helper(problem, &best_medicine, organ);
    
    // Check if we already have this medicine in the body
    let current_contribution = estimate_current_medicine_contribution_heuristic_helper(
        problem,
        state,
        &best_medicine,
        organ,
        property,
    );
    
    let adjusted_deficit = (deficit - current_contribution).max(0.0);
    let adjusted_doses_needed = (adjusted_deficit / peak_effect).ceil().max(1.0);
    
    // Time estimate: time to peak + additional doses (spaced by elimination time / 2 for safety)
    let elimination_time = *problem.static_info.decay_times.get(&best_medicine).unwrap_or(&10);
    let spacing = (elimination_time as f64 / 2.0).max(1.0);
    
    time_to_peak + (adjusted_doses_needed - 1.0) * spacing
}

/// Find alternative medicine if the best one is exhausted
fn find_alternative_medicine_time_heuristic_helper(
    problem: &MedicalProblem,
    state: &State,
    organ: &String,
    property: &String,
    deficit: f64,
    excluded_medicine: &String,
) -> f64 {
    let mut best_time = 1000.0_f64;
    
    for medicine in &problem.static_info.medicines {
        if medicine == excluded_medicine {
            continue;
        }
        
        let doses_applied = state.medicine_doses_applied.get(medicine).unwrap_or(&0);
        let max_doses = problem.static_info.usage_constraints.get(medicine).unwrap_or(&i32::MAX);
        
        if doses_applied >= max_doses {
            continue;
        }
        
        let key = (medicine.clone(), organ.clone(), property.clone());
        let emax_val = problem.static_info.emax.get(&key).unwrap_or(&0.0);
        
        if *emax_val <= 0.0 {
            continue;
        }
        
        let max_dosage = problem.static_info.dosage_sizes
            .get(medicine)
            .and_then(|sizes| sizes.iter().max())
            .unwrap_or(&1);
        
        let ec50_val = problem.static_info.ec50.get(&key).unwrap_or(&1.0);
        let peak_effect = emax_val * (*max_dosage as f64 / (*max_dosage as f64 + ec50_val));
        
        if peak_effect > 0.0 {
            let doses_needed = (deficit / peak_effect).ceil();
            let time_to_peak = estimate_time_to_peak_heuristic_helper(problem, medicine, organ);
            let elimination_time = *problem.static_info.decay_times.get(medicine).unwrap_or(&10);
            let spacing = (elimination_time as f64 / 2.0).max(1.0);
            
            let time_estimate = time_to_peak + (doses_needed - 1.0) * spacing;
            best_time = best_time.min(time_estimate);
        }
    }
    
    best_time
}

/// Estimate time to peak effect based on PK profile
fn estimate_time_to_peak_heuristic_helper(
    problem: &MedicalProblem,
    medicine: &String,
    organ: &String,
) -> f64 {
    if let Some(organs_map) = problem.static_info.graph.get(medicine) {
        if let Some(pk_profile) = organs_map.get(organ) {
            // Find the index of maximum concentration
            let mut max_idx = 0;
            let mut max_val = 0.0_f64;
            
            for (idx, &val) in pk_profile.iter().enumerate() {
                if val > max_val {
                    max_val = val;
                    max_idx = idx;
                }
            }
            
            return max_idx as f64;
        }
    }
    
    // Default: assume peak at 1/3 of elimination time
    let elimination_time = *problem.static_info.decay_times.get(medicine).unwrap_or(&10);
    (elimination_time as f64 / 3.0).max(1.0)
}

/// Estimate current contribution of a medicine already in the body
fn estimate_current_medicine_contribution_heuristic_helper(
    problem: &MedicalProblem,
    state: &State,
    medicine: &String,
    organ: &String,
    property: &String,
) -> f64 {
    // Calculate current PK level for this medicine in this organ
    let mut pk_level = 0.0_f64;
    
    let elimination_time = *problem.static_info.decay_times.get(medicine).unwrap_or(&0);
    
    if let Some(organs_map) = problem.static_info.graph.get(medicine) {
        if let Some(pk_profile) = organs_map.get(organ) {
            if let Some(doses) = state.medicine_history.get(medicine) {
                for &(dose_time, dosage) in doses {
                    let elapsed = state.timestamp - dose_time;
                    if elapsed >= 0 && elapsed < elimination_time && (elapsed as usize) < pk_profile.len() {
                        let factor = pk_profile[elapsed as usize];
                        pk_level += (dosage as f64) * factor;
                    }
                }
            }
        }
    }
    
    // Calculate PD effect
    let key = (medicine.clone(), organ.clone(), property.clone());
    let ec50_val = problem.static_info.ec50.get(&key).unwrap_or(&1.0);
    let emax_val = problem.static_info.emax.get(&key).unwrap_or(&0.0);
    
    if pk_level > 0.0 {
        emax_val * (pk_level / (pk_level + ec50_val))
    } else {
        0.0
    }
}

/// Compute penalty for being close to safety boundaries
fn compute_safety_penalty_heuristic_helper(problem: &MedicalProblem, state: &State) -> f64 {
    let mut penalty = 0.0_f64;
    
    for ((organ, property), value) in &state.organ_properties {
        if let Some((min, max)) = problem.static_info.property_constraints.get(&(organ.clone(), property.clone())) {
            let range = max - min;
            if range <= 0.0 {
                continue;
            }
            
            // Distance from boundaries as fraction of range
            let dist_from_min = (value.0 - min) / range;
            let dist_from_max = (max - value.0) / range;
            
            // Add penalty if we're in the outer 20% of the safe range
            let safety_margin = 0.2;
            
            if dist_from_min < safety_margin {
                penalty += (safety_margin - dist_from_min) * 5.0;
            }
            
            if dist_from_max < safety_margin {
                penalty += (safety_margin - dist_from_max) * 5.0;
            }
        }
    }
    
    penalty
}

\end{lstlisting}

\bibliography{aaai2026,aaaivar}